\documentclass[letterpaper, 10pt, times, preprintl]{elsarticle}
\journal{arXiv}

\usepackage[english]{babel}
\usepackage{graphicx,subcaption} 
\usepackage{multirow}
\usepackage[table,xcdraw]{xcolor}
\usepackage{booktabs}
\usepackage{array}
\usepackage{comment}
\usepackage{float}
\usepackage{pgffor} 

\usepackage{float}

\usepackage{graphicx}
\usepackage[percent]{overpic}
\usepackage{subcaption}
\usepackage[percent]{overpic}    
\usepackage{subcaption}

\usepackage[letterpaper,top=2cm,bottom=2cm,left=3cm,right=3cm,marginparwidth=1.75cm]{geometry}

\usepackage{soul}
\usepackage{amsmath}
\usepackage{amssymb}

\usepackage[numbers]{natbib}

\usepackage{graphicx}
\usepackage[colorlinks=true, allcolors=blue]{hyperref}

\begin{document}

\begin{frontmatter}
\title{Accurate and Uncertainty-Aware Multi-Task Prediction of HEA Properties Using Prior-Guided Deep Gaussian Processes}
\author{Sk Md Ahnaf Akif Alvi$^{a,e}$} 
\corref{mycorrespondingauthor}
\ead{ahnafalvi@tamu.edu}
\author{Mrinalini Mulukutla$^{a}$}
\author{Nicolás Flores$^{a}$}
\author{Danial Khatamsaz$^{a}$}

\author{Jan Janssen$^{d}$}

\author{Danny Perez$^{e}$}
\author{Douglas Allaire$^{b}$}
\author{Vahid Attari$^{a}$}
\author{Raymundo Arróyave$^{a,b,c}$}
    
\address{$^a$Department of Materials Science and Engineering, Texas A\&M University, College Station, TX, USA 77843}

\address{$^b$ J. Mike Walker '66 Department of Mechanical Engineering, Texas A\&M University, College Station, TX, USA 77843}
\address{$^c$ Wm Michael Barnes '64 Department of Industrial and Systems Engineering, Texas A\&M University, College Station, TX, USA 77843}
\address{$^d$ Max-Planck-Institute for Sustainable Materials, D\"usseldorf, Germany, 40237}
\address{$^e$ Theoretical Division T-1, Los Alamos National Laboratory, Los Alamos, NM, USA 87544}

\begin{abstract}

Surrogate modeling techniques have become indispensable in accelerating the discovery and optimization of high-entropy alloys (HEAs), especially when integrating computational predictions with sparse experimental observations. This study systematically evaluates the fitting performance of four prominent surrogate models— conventional Gaussian Processes (cGP), Deep Gaussian Processes(DGP), encoder-decoder neural networks for multi-output regression and XGBoost—applied to a hybrid dataset of experimental and computational properties in the AlCoCrCuFeMnNiV HEA system. We specifically assess their capabilities in predicting correlated material properties, including yield strength, hardness, modulus, ultimate tensile strength, elongation, and average hardness under dynamic/quasi-static conditions, alongside auxiliary computational properties. The comparison highlights the strengths of hierarchical and deep modeling approaches in handling heteroscedastic, heterotopic, and incomplete data commonly encountered in materials informatics. Our findings illustrate that DGP infused with machine learning-based prior  outperform other surrogates by effectively capturing inter-property correlations and input-dependent uncertainty. This enhanced predictive accuracy positions advanced surrogate models as powerful tools for robust and data-efficient materials design.
\end{abstract}

\begin{keyword}
Deep Gaussian Processes
Multi-task Gaussian Processes
High Entropy Alloys
\end{keyword}

\end{frontmatter}

\section{Introduction} 


High-entropy alloys (HEAs) are multicomponent alloys with five or more principal elements in high concentrations that offer a vast compositional design space and often exhibit exceptional mechanical properties \cite{miracle2017critical,george2019high}. The exploration of this space is challenged by limited experimental data and complex composition–property relationships \cite{miracle2017critical,elkatatny2024machine}. Accordingly, data-driven surrogate models have become invaluable in accelerating HEA discovery by predicting material properties by composition, and these challenges have motivated the development and application of various surrogate modeling techniques in recent studies \cite{elkatatny2024machine,wen2019machine,wang2023neural,gao2024data}.

Surrogate models serve as computationally efficient approximations of complex structure–property relationships and are widely used to accelerate materials discovery. Gaussian Processes (GPs) provide calibrated uncertainty estimates, making them effective for sparse, high-fidelity datasets \cite{rasmussen2006gaussian}, while multi-task extensions and co-kriging allow for correlated output modeling \cite{bonilla2008multi}. Deep Gaussian Processes (DGPs) extend this capability to hierarchical, nonlinear functions \cite{damianou2013deep}, offering an advantage in capturing complex material behavior. In contrast, tree-based methods like XGBoost~\cite{chen2016xgboost} or encoder-decoder mapping of inputs to outputs~\cite{attari2024decoding} are often easier to scale and tune but lack native uncertainty quantification unless modified. As a result, model selection should be guided by data availability, dimensionality, and the need for uncertainty modeling in the optimization loop. Hybrid modeling strategies that combine the strengths of neural networks (for representation learning) and probabilistic models (for uncertainty quantificdaation) offer a promising path forward—leveraging expressive power while retaining decision-making confidence in sparse-data regimes.

Recent studies have combined machine learning with high-throughput calculations and experiments to identify novel alloys with targeted properties\cite{rao2022machine}. In particular, Bayesian optimization (BO) frameworks for materials design rely on accurate surrogate models to guide experiments toward optimal candidates\cite{wang2020multi}. In our previous work, we addressed the discovery of HEAs with specific properties by using advanced Gaussian process models (conventional Gaussian process, multitask Gaussian process and deep Gaussian process) within a BO loop\cite{alvi2025hierarchical}. We also explored deep Gaussian process (DGP) surrogates in that context, which – along with MTGPs – were able to capture the coupled trends in bulk modulus and coefficient of thermal expansion, while also accelerating the search for promising alloy compositions in Fe-Cr-Ni-Co-Cu HEA space. In particular, Khatamsaz \textit{ et al.}\cite{khatamsaz2023multi} employed a multitask Gaussian process (MTGP) approach to jointly model the yield strength, the Pugh ratio and the Cauchy pressure in a simulated Mo-Ti-Nb–V–W alloy system, enabling efficient multiobjective optimization for high strength and ductility.

Building on the successes of these optimization-focused studies, we now shift our attention toward a detailed evaluation of surrogate modeling techniques. Moving beyond optimization, the present work assesses how well different models fit a hybrid data set consisting of both experimental measurements and computational predictions. In particular, we compare multiple modeling approaches, conventional single-layer Gaussian processes (cGPs), DGPs, a custom encoder–decoder neural network for multi-output regression, and XGBoost, on their ability to learn the composition–property relationships in the Al–Co–Cr–Cu–Fe–Mn–Ni–V HEA system. DGPs are a hierarchical extension of GPs that can capture complex, nonlinear mappings by composing multiple Gaussian process layers\cite{damianou2013deep}, and we hypothesize that this added depth enables DGPs to model heteroscedastic and nonstationary behavior often observed in materials data better than a standard cGP. The encoder–decoder model, on the other hand, represents a deterministic deep learning approach: it “encodes” alloy compositions into a learned feature representation and “decodes” this latent vector to simultaneously predict multiple properties\cite{ban2023effective}.

As a baseline, we include a conventional XGBoost model for each property to illustrate the limitations of ignoring inter-property correlations and data uncertainty. The material data set used in this study spans the AlCoCrCuFeMnNiV composition space, an 8-element HEA system of interest for its high configurational entropy and the potential for exceptional mechanical performance. This data set is assembled from disparate sources, having experimentally calculated properties as 'main tasks' ans some additional computationally estimated properties as 'auxiliary tasks'. The inclusion of these additional 'tasks' provides additional information that can be used by multiple output models to improve predictions of the main experimental properties, analogous to multifidelity or multisource learning in other contexts\cite{khatamsaz2021efficiently, ghoreishi2019efficient}.

Given the diverse origins of this dataset, not every alloy composition has a complete set of measurements; some samples have missing experimental values, while corresponding computational predictions may exist, and vise versa. This reflects a common situation in material informatics, where gathering a complete set of property data for every sample is infeasible, and one must utilize incomplete and noisy information from various sources. Consequently, the challenges posed by this HEA dataset underscore the need for surrogate models capable of handling correlated output, heteroscedastic uncertainties, and missing data. Moreover, many of the mechanical properties in HEAs are interdependent---for example, hardness and yield strength are often correlated since they both relate to underlying strengthening mechanisms. A multi-output model can exploit such correlations to improve the prediction accuracy for each task, especially when one property has abundant data and another is data sparse\cite{bonilla2008multi, ban2023effective}.

In addition to the challenges of missing data, noise levels and predictive difficulties can vary between properties. Experimental measurements like elongation may have higher uncertainty or variability than, say, VEC (which is deterministically computed), leading to heteroscedastic behavior. The sources and treatment of uncertainty differ across the various properties considered in this work. For experimentally measured properties—such as hardness, modulus, or elongation—uncertainty is quantified using the standard deviation across three replicate measurements per alloy, capturing variability due to factors like local microstructural differences and measurement noise. These properties can exhibit heteroscedasticity across the dataset as their variance may depend on composition or processing conditions. In contrast, some descriptors used in this study are computed rather than measured, and have different uncertainty profiles. For example, valence electron concentration (VEC) is computed deterministically using a rule-of-mixtures approach and currently does not include an associated uncertainty estimate in our framework. However, we recognize that VEC, like any empirical descriptor, has model-form uncertainty based on its simplifying assumptions. Similarly, stacking fault energy (SFE) values used in this study were generated using a machine learning model trained on density functional theory (DFT) data, as described by Khan et al.\cite{khan_towards_2022}, while yield strength estimates were computed using the Curtin–Varvenne solid solution strengthening model \cite{varvenne2016theory}. Both of these computed descriptors inherit uncertainty from their underlying models and training data, although explicit uncertainty quantification was not included. To support future use and analysis, we plan to release uncertainty estimates alongside the dataset—including statistical uncertainties for experimental data and model-based estimates for computed properties—to enable more robust uncertainty-aware modeling and optimization across the HEA design space. Traditional single-task GPs can struggle in this setting because they assume a single noise variance per task and cannot borrow strength from related tasks. In contrast, co-regionalized DGP frameworks can naturally handle task-specific noise and allow information transfer across properties\cite{alvarez2012kernels}. Likewise, DGPs can effectively model input-dependent noise by virtue of their layered construction – the output of one GP layer (with its own variance) feeds into the next, allowing the model to represent variability that changes with composition\cite{salimbeni2017doubly}. Neural networks can also accommodate heteroscedasticity by learning complex nonlinear interactions, though they typically require larger datasets to generalize well and often treat noise implicitly (e.g., through regularization techniques). Importantly, both the DGP and the encoder-decoder model can handle missing outputs during training, as they are trained with a loss (or likelihood) that includes only observed data for each task, enabling the use of all available partial information.

Despite the advances in surrogate modeling and optimization for HEAs, a notable gap remains: while numerous studies have demonstrated the successful application of individual surrogate models for property prediction and optimization, a robust, non-parametric, multi-property model being able to predict uncertainty as well on a unified, multi-property alloy dataset is lacking\cite{elkatatny2024machine,wen2019machine,wang2023neural,gao2024data}. In this paper, our aim is to fill this gap by systematically benchmarking DGPs, DGPs with encoder-decoder prior, encoder–decoder multi-output architectures, cGPs, and XGBoost on the BIRDSHOT dataset. Through this comparison, we evaluate the ability of each method to fit the complex property landscape of HEAs and faithfully capture property correlations and uncertainties. The insights from this study provide guidance on selecting surrogate models for future materials informatics efforts, especially in scenarios with sparse, noisy, and multisource data. We show that appropriately accounting for task correlations and data heterogeneity – via advanced models like DGPs – can markedly improve predictive performance, thereby enabling more reliable downstream tasks such as Bayesian alloy design and multiobjective optimization.

\section{Results}

\subsection{BIRDSHOT Data}

\begin{figure}[htbp]
    \centering
    \includegraphics[width=\textwidth]{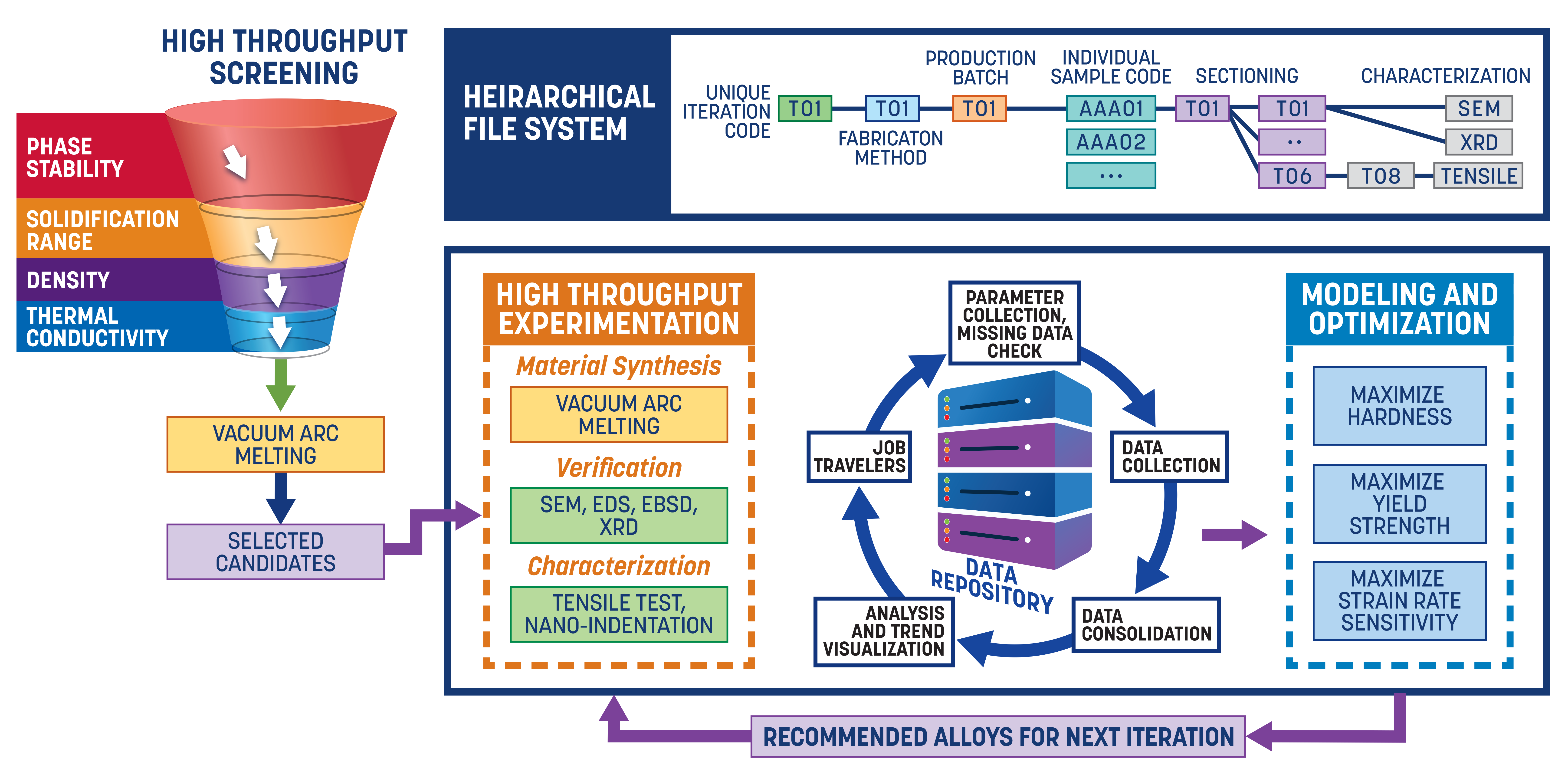} 
    \caption{Schematic overview of the BIRDSHOT data acquisition workflow. The dataset results from a multi-institutional effort combining alloy synthesis, mechanical testing across various strain rates (including nanoindentation, strain rate jump tests, small punch tests, and SHPB), and structural and compositional verification (EBSD, XRD, SEM-EDS). This integrated workflow ensures fidelity and consistency across the high-dimensional alloy space explored.}
    \label{fig:birdshot_workflow}
\end{figure}

The BIRDSHOT dataset is a comprehensive, high-fidelity collection of mechanical, structural, and compositional data covering over 100 distinct high-entropy alloy (HEA) compositions in the Al–V–Cr–Mn–Fe–Co–Ni–Cu system~\cite{hastings2024interoperable,mulukutla2024illustrating}---see Fig.~\ref{fig:birdshot_workflow}. It is the product of two complementary experimental campaigns---the first campaign has already been thoroughly described in~\cite{hastings2024interoperable}---each targeting slightly different regions of the compositional space and emphasizing different mechanical performance metrics. The first campaign focused on a six-element FCC HEA system (Al–V–Cr–Fe–Co–Ni), while the second campaign expanded the design space to include Mn and Cu. Together, these efforts explored a filtered design space consisting of tens of thousands of feasible alloys, ultimately selecting and synthesizing more than 100 unique compositions. These alloys were chosen to span a broad and representative portion of the chemically complex HEA landscape, under multiple design constraints related to phase stability, thermal, and mechanical properties.

Alloys in the dataset were synthesized using vacuum arc melting (VAM) under strictly controlled and standardized protocols\cite{hastings2024interoperable}. Each 30--35~g ingot was produced from high-purity elements, flipped and remelted multiple times to ensure chemical homogeneity, and subjected to homogenization followed by mechanical working (hot forging or cold rolling and recrystallization). These protocols were applied uniformly across both campaigns, ensuring consistency in processing history. Compositional accuracy was confirmed via SEM-EDS with average deviations from targets below 1\%, while structural phase verification was carried out using XRD and EBSD. This uniformity enables direct comparison of properties across compositions without the confounding effects of processing variability.

Each alloy underwent a suite of structural and mechanical characterizations, with properties measured from subsamples taken from the \emph{same physical ingot}. This eliminates ambiguity often introduced when comparing properties measured on nominally identical but independently prepared samples. As a result, the dataset provides true cross-property, multimodal measurements—such as yield strength, ultimate tensile strength, hardness, strain-rate sensitivity, and modulus—directly coupled to each alloy’s verified composition and microstructure.

Mechanical behavior was characterized across a wide range of strain rates. Miniaturized tensile tests were conducted to extract yield and ultimate strengths and strain hardening ratios. High-strain-rate nanoindentation, including strain rate jump tests, was used to determine rate-sensitive hardness and modulus. Dynamic performance was further evaluated via small punch tests, split Hopkinson pressure bar (SHPB) compression testing, and Laser-Induced Projectile Impact Testing (LIPIT), enabling mechanical assessment from $10^{-4}$ to $10^7$~s$^{-1}$. Microstructural analyses using SEM, EBSD, and XRD verified single-phase FCC stability and ruled out the presence of deleterious secondary phases.

A key innovation of the BIRDSHOT effort lies in its use of a Bayesian discovery framework\cite{arroyave2022perspective} to guide experimental exploration. This closed-loop approach integrates machine learning, physical modeling, and experimental data to iteratively select the most informative alloy compositions. At each iteration, a surrogate model—typically a Gaussian Process (GP)—is trained on prior experimental data to predict material properties across the compositional space, along with associated uncertainty. An acquisition function, such as Expected Hypervolume Improvement (EHVI), is used to rank candidate alloys by their potential to improve the Pareto front defined by multiple objectives. A batch of promising compositions is selected, synthesized, tested, and then fed back into the model for the next iteration. This strategy balances exploration of under-sampled regions with exploitation of high-performing areas and led to the discovery of a robust multi-objective Pareto set while sampling less than 0.2\% of the filtered design space~\cite{hastings2024interoperable}.

The two campaigns established a rigorously curated, multi-modal, and multi-objective dataset that not only provides broad coverage of a chemically rich compositional space, but also captures cross-property correlations at an unprecedented level of fidelity. This makes the BIRDSHOT dataset a valuable resource for both mechanistic understanding and data-driven modeling of complex alloy systems---in this specific case, the FCC HEA space.

In addition to experimental data, the dataset includes a suite of simulation-derived descriptors computed for each alloy composition. Valence electron concentration (VEC) was calculated using a rule-of-mixtures approximation, based on atomic fractions of constituent elements. Yield strength estimates were obtained using the Curtin–Varvenne model \cite{varvenne2016theory}, which computes solid solution strengthening based on atomic size and modulus mismatch among elements. Stacking fault energy (SFE) values were predicted using a machine learning model trained on density functional theory (DFT) data, developed by Khan et al.~\cite{khan_towards_2022}, providing composition-sensitive estimates of SFE across the HEA design space. 

Depth of penetration, representing a measure of ballistic resistance, was predicted through finite element simulations using ABAQUS. These simulations modeled the impact of a spherical alumina projectile onto a cylindrical HEA target traveling at 500 m/s. Each alloy was modeled using a Cowper–Symonds viscoplasticity framework, calibrated with experimental stress-strain data from quasi-static tensile and Split-Hopkinson Pressure Bar (SHPB) experiments.  While descriptors such as VEC, SFE, and predicted yield strength are available for all alloy compositions in the dataset, depth of penetration is currently available for a subset of alloys tested for SHPB tests.

\subsection{Effect of Elemental Composition on Experimental Properties}

\begin{figure}[H]
    \centering
    \includegraphics[width=0.8\textwidth]{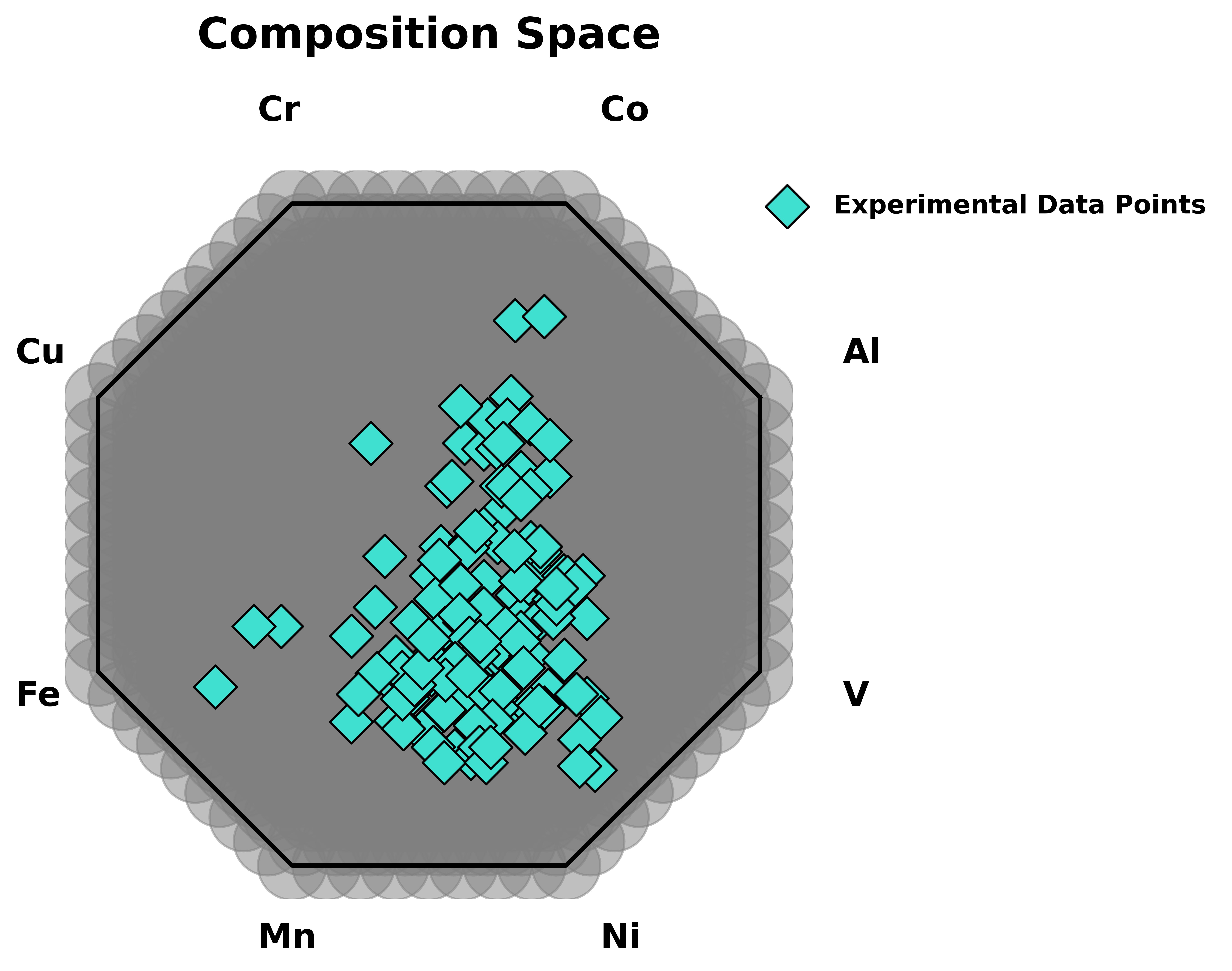}
    \caption{Barycentric Plot of available experimental data points in  Al–V–Cr–Mn–Fe–Co–Ni–Cu composition space.}
    \label{fig:bary}
\end{figure}

The experimental data set contains data points scattered throughout the composition space. The selection strategy of the data points is based on BO of two campaigns, as mentioned in the method section. The available data points are illustrated in a barycentric plot of the  Al–V–Cr–Mn–Fe–Co–Ni–Cu space in figure \ref{fig:bary}.

\begin{figure}[H]
    \centering
    \includegraphics[width=0.9\textwidth]{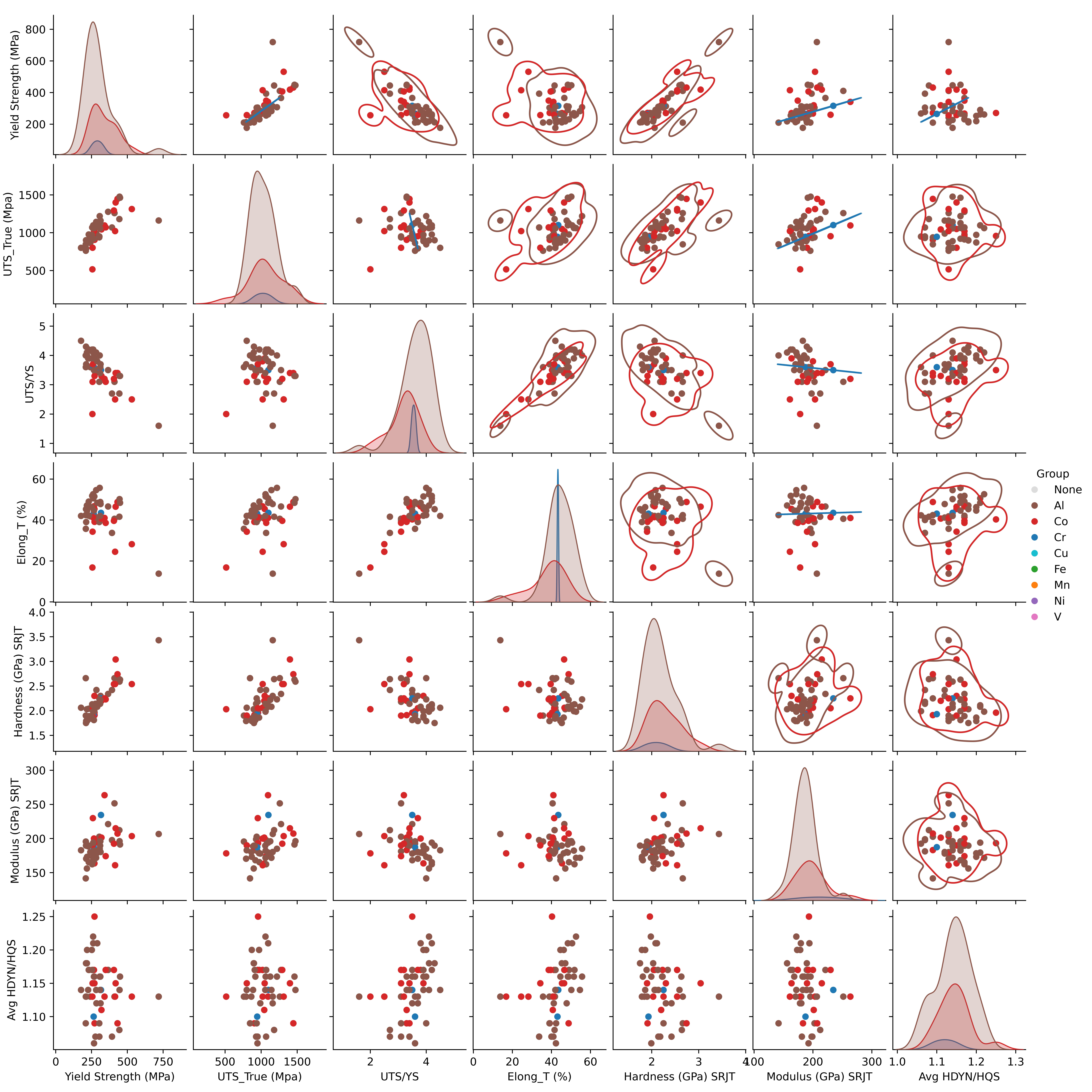}
    \caption{Pairwise plot illustrating correlations between different experimental properties.}
    \label{fig:Pairwise}
\end{figure}

To further investigate correlations among properties, we conducted pairwise plots (Figure~\ref{fig:Pairwise}), Our goal is to demonstrate that considering correlation among the properties yields a better fit for the surrogates. The pairplot reveal strong positive correlations between yield strength, hardness, and UTS, emphasizing their interdependent nature. Ductility (Elong\_T) shows a clear inverse relationship with these strength-related properties, highlighting the common strength-ductility trade-off observed in high-entropy alloys \cite{miracle2017critical}.

These visual analyses establish the foundational understanding of the underlying correlations and elemental influences critical to interpreting the surrogate modeling performance discussed subsequently.

\subsection{Comparison of Model Performances}

In this study, we systematically evaluated seven distinct surrogate modeling approaches to accurately predict material properties within the AlCoCrCuFeMnNiV HEA space. The methods assessed include: (i) Deep Gaussian Processes (DGP) incorporating priors trained on all tasks (ii) DGP incorporating priors trained on only main objectives, (iii) DGP without priors utilizing all tasks, (iv) DGP without priors trained solely on main objectives, (v) conventional Gaussian Processes (GP) with no correlated kernel, (vi) XGBoost and (Vii) encoder-decoder model used for regression tasks~\cite{attari2024decoding}.

\begin{table}[!ht]
\centering
\caption{Summary of surrogate modeling strategies evaluated in this study.}
\small
\label{tab:model_summary}
\begin{tabular}{p{3cm}llp{7cm}}
\hline
\textbf{Model Name} & \textbf{Uses Prior} & \textbf{Tasks Included} & \textbf{Description} \\
\hline
HDGP P-All & Yes  & Main + Auxiliary & DGP with priors derived from an encoder-decoder model trained on main tasks. \\
HDGP P-Main & Yes  & Main Only & DGP with priors from an encoder-decoder model trained only on main tasks. \\
HDGP NP-All & No  & Main + Auxiliary & DGP without using prior knowledge.  \\
HDGP NP-Main & No  & Main Only & DGP without priors or auxiliary properties. \\
GP (no corr. kernel) & No  & Individual & Conventional GP trained independently for each property. Does not model inter-property correlations. \\
XGBoost & No  & Per Task & Gradient-boosted decision tree model applied separately to each property (classical regression baseline). \\
Encoder-Decoder (regularized dense network) & No  & Main only & Multi-target regression model based on decoding nonlinearities in data (Also used as a prior in HDGP configuration). \\
\hline
\end{tabular}
\end{table}

A unique aspect of our modeling approach involves incorporating prior knowledge derived from a previously trained encoder-decoder neural regression model. We fixed the train-test split between the two models(DGP and encoder-decoder) for the cases where DGPs use encoder-decoder as prior. This prevents \textbf{data leakage}. By preventing data leakage, we ensure our test set is totally unseen by the DGP models during training, as it does not exist in the training set of encoder-decoder used by DGP as prior.  

This prior was injected into the DGP models by subtracting predicted prior values from the training data outputs, thereby focusing the DGP on modeling residuals. After the prediction phase, the priors were re-added to generate the final predictions, which were then visualized and evaluated. This technique leverages existing domain knowledge and has the potential to significantly enhance predictive performance, especially in sparse and noisy data environments \cite{ban2023effective, vela2023data}.

\begin{figure}[!ht]
    \centering
    \includegraphics[width=1\textwidth]{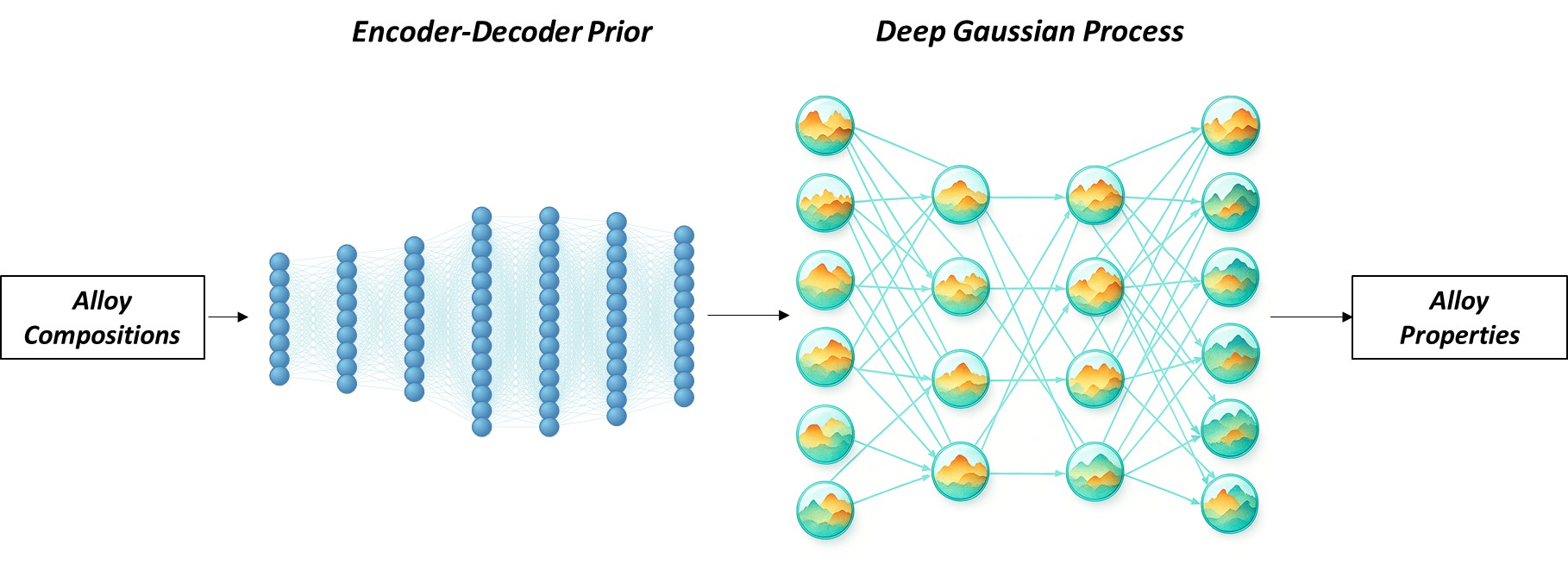}
    \caption{Schematic representation of the two-layered variational Deep Gaussian Process (DGP) architecture with Encoder-Decoder based prior injection. Each variational Multi-task Gaussian Process (MTGP) layer reduces dimensionality from input to latent representation, governed by the reduction parameter.}
    \label{fig:dgp_architecture}
\end{figure}

\begin{figure}[htbp]
  \centering
  \foreach \i/\lab in {
    0/(a),1/(b),2/(c),
    3/(d),4/(e),5/(f),
    6/(g),7/(h),8/(i)
  }{%
    \begin{overpic}[width=0.32\textwidth,
                    height=0.24\textheight,
                    keepaspectratio]{task_\i_%
      \ifnum\i=0 VarvYS_pred_MPa__at_298K_split_2b.png\fi
      \ifnum\i=1 SFE_calc_split_2b.png\fi
      \ifnum\i=2 VEC_Avg_split_2b.png\fi
      \ifnum\i=3 Yield_Strength__MPa__split_2b.png\fi
      \ifnum\i=4 UTS_True__Mpa__split_2b.png\fi
      \ifnum\i=5 UTS_YS_split_2b.png\fi
      \ifnum\i=6 Elong_T_____split_2b.png\fi
      \ifnum\i=7 Hardness__GPa__SRJT_split_2b.png\fi
      \ifnum\i=8 Modulus__GPa__SRJT_split_2b.png\fi
    }
      \put(2,95){\large\bfseries \lab}
    \end{overpic}%
    \ifnum\i<2 \hfill\fi
    \ifnum\i=2 \\[0.5cm]\fi
    \ifnum\i=5 \\[0.5cm]\fi
  }

  \begin{overpic}[width=0.32\textwidth,
                  height=0.24\textheight,
                  keepaspectratio]{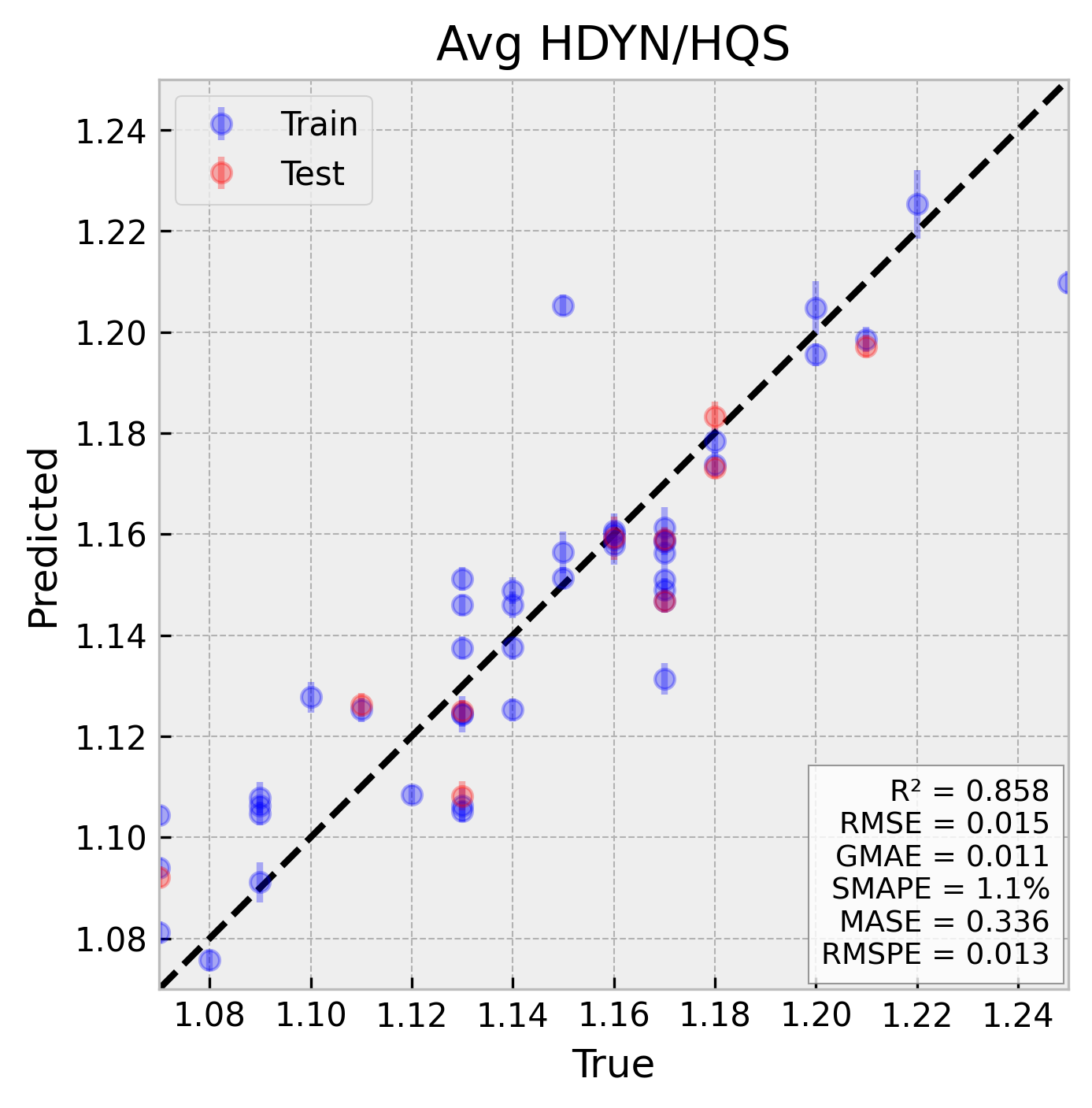}
    \put(2,95){\large\bfseries (j)}
  \end{overpic}\hfill
  \begin{overpic}[width=0.32\textwidth,
                  height=0.24\textheight,
                  keepaspectratio]{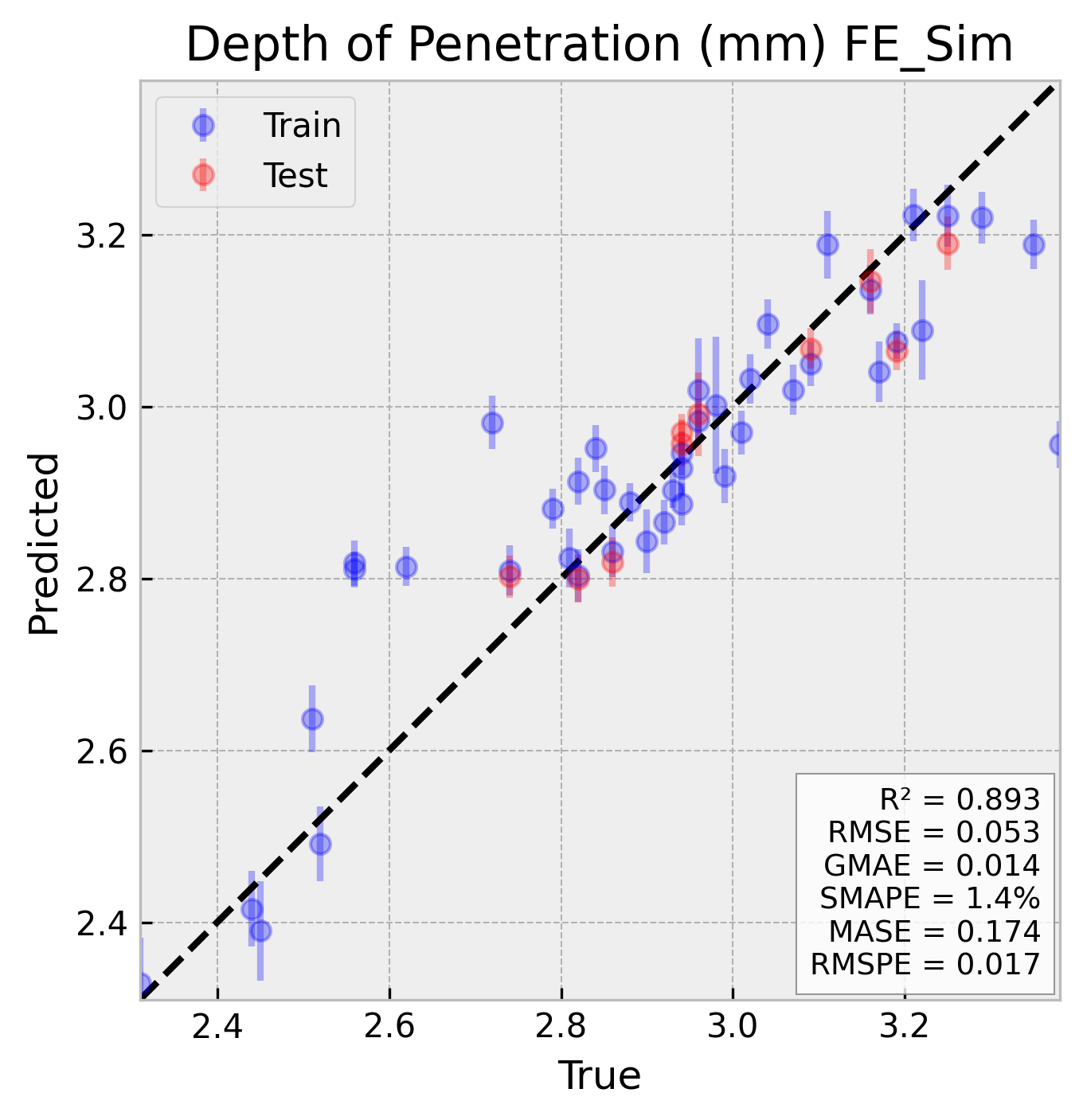}
    \put(2,95){\large\bfseries (k)}
  \end{overpic}

  \caption{HDGP-P-All model results:  
    (a) Predicted VarvYS at 298 K; (b) SFE calculation; (c) Average VEC;  
    (d) Yield Strength (MPa); (e) True UTS (MPa); (f) UTS/YS Ratio;  
    (g) Elongation; (h) Hardness (GPa); (i) Modulus (GPa);  
    (j) Average HDYN HQS; (k) Depth of Penetration (mm).}
  \label{fig:alltasks_grid}
\end{figure}

To comprehensively evaluate model performance, we utilized multiple standard regression metrics, including the coefficient of determination (\(R^2\)), Root Mean Squared Error (RMSE), Symmetric Mean Absolute Percentage Error (SMAPE), and Mean Averaged Error (MAE). Additionally, Spearman's rank correlation coefficient was computed to assess the consistency of predicted rankings relative to actual data, providing insights into the models' abilities to preserve order, which is critical for practical materials selection scenarios.

The final reported metrics were obtained by averaging results from five different randomized train-test splits to ensure robust and reliable performance evaluation. Specifically, an 80-20 train-test split was employed for the DGP, GP, XGBoost models, and encoder-decoder model. For the DGP models, the optimal reduction parameter identified during the hyperparameter tuning stage was consistently applied across these evaluations.

In the following section, we will provide detailed comparisons and discussions of the performance metrics across the seven evaluated surrogate models, elucidating the strengths, limitations, and practical implications of each approach for multi-task prediction in high-entropy alloy datasets.

\begin{table}[H]
\centering
\footnotesize
\caption{Test performance metrics (Tasks 1--6). Best values (lowest RMSE, MAE, SMAPE; highest R$^2$ and Spearman) are both bold and underlined.}
\label{tab:consolidated_metrics_1}
\resizebox{\columnwidth}{!}{%
\begin{tabular}{|l|l|c|c|c|c|c|c|c|}
\hline
\textbf{Task} & \textbf{Metric} & \textbf{XGB} & \textbf{cGP} & \textbf{HDGP NP-All} & \textbf{HDGP NP-Main} & \textbf{HDGP P-All} & \textbf{HDGP P-Main} & \textbf{EncDec} \\ \hline
VarvYS pred(MPa) at 298K & RMSE & \underline{\textbf{63.079 $\pm$ 43.580}} & 85.254 $\pm$ 60.183 & 126.506 $\pm$ 51.467 & -- & 133.2247 $\pm$ 25.0982 & -- & -- \\ 
 & MAE & 34.404 $\pm$ 21.687 & \underline{\textbf{32.621 $\pm$ 26.955}} & 89.187 $\pm$ 31.297 & -- & 90.7540 $\pm$ 13.8952 & -- & -- \\ 
 & R2 & \underline{\textbf{0.852 $\pm$ 0.122}} & 0.712 $\pm$ 0.201 & 0.405 $\pm$ 0.164 & -- & 0.4084 $\pm$ 0.1114 & -- & -- \\ 
 & Spearman & \underline{\textbf{0.944 $\pm$ 0.029}} & 0.898 $\pm$ 0.069 & 0.667 $\pm$ 0.146 & -- & 0.7131 $\pm$ 0.0917 & -- & -- \\ 
 & SMAPE & 8.6\% $\pm$ 4.0\% & \underline{\textbf{6.8\% $\pm$ 5.2\%}} & 25.2\% $\pm$ 6.2\% & -- & 27.75\% $\pm$ 2.95\% & -- & -- \\ \hline
SFE calc & RMSE & 160.015 $\pm$ 31.749 & 154.823 $\pm$ 34.594 & 144.515 $\pm$ 29.097 & -- & \underline{\textbf{141.9662 $\pm$ 12.8200}} & -- & -- \\ 
 & MAE & 103.249 $\pm$ 11.891 & 101.121 $\pm$ 17.958 & 94.588 $\pm$ 11.523 & -- & \underline{\textbf{94.5297 $\pm$ 11.6745}} & -- & -- \\ 
 & R2 & 0.126 $\pm$ 0.126 & 0.186 $\pm$ 0.139 & \underline{\textbf{0.289 $\pm$ 0.100}} & -- & 0.1921 $\pm$ 0.1753 & -- & -- \\ 
 & Spearman & \underline{\textbf{0.686 $\pm$ 0.066}} & 0.567 $\pm$ 0.099 & 0.677 $\pm$ 0.117 & -- & 0.6124 $\pm$ 0.1351 & -- & -- \\ 
 & SMAPE & \underline{\textbf{72.3\% $\pm$ 8.6\%}} & 78.0\% $\pm$ 10.2\% & 76.3\% $\pm$ 5.6\% & -- & 82.61\% $\pm$ 15.56\% & -- & -- \\ \hline
VEC Avg & RMSE & 13.345 $\pm$ 2.473 & \underline{\textbf{0.904 $\pm$ 0.172}} & 62.739 $\pm$ 15.508 & -- &  61.4221 $\pm$ 23.8710 & -- & -- \\ 
 & MAE & 9.900 $\pm$ 1.613 & \underline{\textbf{0.559 $\pm$ 0.122}} & 45.720 $\pm$ 10.894 & -- & 43.5553 $\pm$ 15.8566 & -- & -- \\ 
 & R2 & 0.986 $\pm$ 0.005 & \underline{\textbf{1.000 $\pm$ 0.000}} & 0.690 $\pm$ 0.146 & -- & 0.6758 $\pm$ 0.1728 & -- & -- \\ 
 & Spearman & 0.992 $\pm$ 0.005 & \underline{\textbf{0.999 $\pm$ 0.000}} & 0.837 $\pm$ 0.073 & -- &  0.7858 $\pm$ 0.1334 & -- & -- \\ 
 & SMAPE & 1.5\% $\pm$ 0.3\% & \underline{\textbf{0.1\% $\pm$ 0.0\%}} & 6.8\% $\pm$ 1.8\% & -- & 6.52\% $\pm$ 2.55\% & -- & -- \\ \hline
Yield Strength (MPa) & RMSE & 89.158 $\pm$ 12.990 & 79.908 $\pm$ 10.563 & 77.999 $\pm$ 7.234 & 63.951 $\pm$ 12.507 &  \underline{\textbf{39.4717 $\pm$ 13.7846}} & 39.8794 $\pm$ 14.8138 & 43.9739 $\pm$ 19.9147 \\ 
 & MAE & 63.459 $\pm$ 9.197 & 57.057 $\pm$ 6.396 & 54.650 $\pm$ 6.106 & 46.235 $\pm$ 9.297 & 28.7471 $\pm$ 11.0498 & \underline{\textbf{28.5152 $\pm$ 12.4962}} &  31.9583 $\pm$ 15.4363 \\ 
 & R2 & 0.585 $\pm$ 0.184 & 0.679 $\pm$ 0.110 & 0.685 $\pm$ 0.115 & 0.732 $\pm$ 0.050 &  \underline{\textbf{0.8558 $\pm$ 0.1359}} & 0.8495 $\pm$ 0.1464 & 0.8038 $\pm$ 0.2305  \\ 
 & Spearman & 0.777 $\pm$ 0.093 & 0.838 $\pm$ 0.062 & 0.874 $\pm$ 0.061 & 0.887 $\pm$ 0.024 & 0.9099 $\pm$ 0.0916 & \underline{\textbf{0.9138 $\pm$ 0.0894}} & 0.8952 $\pm$ 0.1255 \\ 
 & SMAPE & 15.8\% $\pm$ 2.7\% & 14.3\% $\pm$ 1.6\% & 13.7\% $\pm$ 1.9\% & 11.8\% $\pm$ 1.4\% & \underline{\textbf{7.64\% $\pm$ 3.04\%}} & 7.64\% $\pm$ 3.41\% & 8.68\% $\pm$ 4.12\% \\ \hline\%
UTS True (Mpa) & RMSE & 182.564 $\pm$ 28.431 & 149.748 $\pm$ 25.830 & 146.172 $\pm$ 26.465 & 180.835 $\pm$ 55.784 & 103.5612 $\pm$ 42.8069 & \underline{\textbf{98.1523 $\pm$ 38.1771}} & 118.3553 $\pm$ 50.7326  \\ 
 & MAE & 138.201 $\pm$ 22.664 & 110.308 $\pm$ 17.711 & 106.354 $\pm$ 15.304 & 130.595 $\pm$ 42.525 & 74.4103 $\pm$ 32.1561 & \underline{\textbf{70.2329 $\pm$ 28.4956}} & 87.2348 $\pm$ 41.1554  \\ 
 & R2 & 0.291 $\pm$ 0.375 & 0.574 $\pm$ 0.042 & 0.595 $\pm$ 0.046 & 0.349 $\pm$ 0.442 & 0.7786 $\pm$ 0.1402 & \underline{\textbf{0.7996 $\pm$ 0.1224}} & 0.6772 $\pm$ 0.3124 \\ 
 & Spearman & 0.630 $\pm$ 0.085 & 0.760 $\pm$ 0.039 & 0.806 $\pm$ 0.033 & 0.687 $\pm$ 0.204 & 0.8741 $\pm$ 0.0925 & \underline{\textbf{0.8761 $\pm$ 0.0982}} & 0.8085 $\pm$ 0.1940 \\ 
 & SMAPE & 15.6\% $\pm$ 2.9\% & 12.4\% $\pm$ 2.6\% & 12.1\% $\pm$ 2.3\% & 14.9\% $\pm$ 5.1\% & 8.91\% $\pm$ 4.08\% & \underline{\textbf{8.40\% $\pm$ 3.62\%}} & 10.44\% $\pm$ 5.02\% \\ \hline
UTS/YS & RMSE & 0.503 $\pm$ 0.071 & 0.417 $\pm$ 0.052 & 0.364 $\pm$ 0.066 & 0.382 $\pm$ 0.057 & 0.3646 $\pm$ 0.1254 & \underline{\textbf{0.3636 $\pm$ 0.1405}} & 0.525 $\pm$ 0.144 \\ 
 & MAE & 0.383 $\pm$ 0.051 & 0.316 $\pm$ 0.043 & 0.271 $\pm$ 0.034 & 0.294 $\pm$ 0.049 & 0.2458 $\pm$ 0.0752 & \underline{\textbf{0.2447 $\pm$ 0.0881}} & 0.382 $\pm$ 0.108 \\ 
 & R2 & 0.678 $\pm$ 0.088 & 0.785 $\pm$ 0.028 & \underline{\textbf{0.835 $\pm$ 0.039}} & 0.817 $\pm$ 0.065 & 0.8032 $\pm$ 0.1312 & 0.7988 $\pm$ 0.1443 & 0.556 $\pm$ 0.253 \\ 
 & Spearman & 0.778 $\pm$ 0.038 & 0.851 $\pm$ 0.040 & 0.844 $\pm$ 0.055 & 0.878 $\pm$ 0.043 & 0.8856 $\pm$ 0.0748 & \underline{\textbf{0.8886 $\pm$ 0.0797}} & 0.809 $\pm$ 0.093 \\ 
 & SMAPE & 15.8\% $\pm$ 1.6\% & 12.8\% $\pm$ 1.5\% & 11.6\% $\pm$ 1.0\% & 11.8\% $\pm$ 2.1\% & \underline{\textbf{9.93\% $\pm$ 3.43\%}} & 10.06\% $\pm$ 4.08\% & 15.301\%  $\pm$ 4.099\% \\ \hline
\end{tabular}%
}
\end{table}

\begin{table}[H]
\centering
\footnotesize
\caption{Test performance metrics (Tasks 7--11). Best values (lowest RMSE, MAE, SMAPE; highest R$^2$ and Spearman) are both bold and underlined.}
\label{tab:consolidated_metrics_2}
\resizebox{\columnwidth}{!}{%
\begin{tabular}{|l|l|c|c|c|c|c|c|c|}
\hline
\textbf{Task} & \textbf{Metric} & \textbf{XGB} & \textbf{cGP} & \textbf{HDGP NP-All} & \textbf{HDGP NP-Main} & \textbf{HDGP P-All} & \textbf{HDGP P-Main} & \textbf{EncDec} \\ \hline
Elong (\%) & RMSE & 7.782 $\pm$ 1.324 & 7.363 $\pm$ 1.222 & 7.278 $\pm$ 1.449 & 7.776 $\pm$ 2.231 & 6.2017 $\pm$ 3.0606 & \underline{\textbf{6.1778 $\pm$ 3.0963}} & 6.6826 $\pm$ 3.1484 \\ 
 & MAE & 5.635 $\pm$ 1.061 & 5.212 $\pm$ 0.646 & 5.263 $\pm$ 0.757 & 5.757 $\pm$ 1.341 & \underline{\textbf{3.8575 $\pm$ 1.5044}} & 3.9204 $\pm$ 1.6349 & 4.2596 $\pm$ 1.7478 \\ 
 & R2 & 0.338 $\pm$ 0.214 & 0.367 $\pm$ 0.290 & 0.342 $\pm$ 0.457 & 0.562 $\pm$ 0.200 & 0.6909 $\pm$ 0.2418 & \underline{\textbf{0.6915 $\pm$ 0.2434}} & 0.6476 $\pm$ 0.2678 \\ 
 & Spearman & 0.740 $\pm$ 0.055 & 0.725 $\pm$ 0.117 & 0.692 $\pm$ 0.155 & 0.743 $\pm$ 0.096 & \underline{\textbf{0.8413 $\pm$ 0.1270}} & 0.8389 $\pm$ 0.1378 & 0.8044 $\pm$ 0.1501  \\ 
 & SMAPE & 21.3\% $\pm$ 5.7\% & 20.4\% $\pm$ 4.2\% & 20.7\% $\pm$ 5.2\% & 25.2\% $\pm$ 5.9\% & \underline{\textbf{17.85\% $\pm$ 4.71\%}}  & 18.01\% $\pm$ 4.89\% & 19.39\% $\pm$ 6.14\% \\ \hline
Hardness (GPa) SRJT & RMSE & 0.371 $\pm$ 0.096 & 0.291 $\pm$ 0.041 & 0.300 $\pm$ 0.031 & 0.331 $\pm$ 0.064 & \underline{\textbf{0.1677 $\pm$ 0.0465}} & 0.1706 $\pm$ 0.0514 & 0.1851 $\pm$ 0.0554 \\ 
 & MAE & 0.280 $\pm$ 0.068 & 0.216 $\pm$ 0.027 & 0.238 $\pm$ 0.026 & 0.230 $\pm$ 0.041 & \underline{\textbf{0.1284 $\pm$ 0.0367}} & 0.1305 $\pm$ 0.0441 & 0.1366 $\pm$ 0.0411\\ 
 & R2 & 0.514 $\pm$ 0.144 & 0.696 $\pm$ 0.050 & 0.665 $\pm$ 0.096 & 0.561 $\pm$ 0.166 & \underline{\textbf{0.8710 $\pm$ 0.1144}} & 0.8614 $\pm$ 0.1326 & 0.8371 $\pm$ 0.1530  \\ 
 & Spearman & 0.781 $\pm$ 0.041 & 0.831 $\pm$ 0.012 & 0.794 $\pm$ 0.086 & 0.759 $\pm$ 0.078 & 0.8956 $\pm$ 0.0888 & \underline{\textbf{0.8988 $\pm$ 0.0874}} & 0.8846 $\pm$ 0.0931 \\ 
 & SMAPE & 10.6\% $\pm$ 2.3\% & 8.3\% $\pm$ 0.8\% & 9.3\% $\pm$ 1.2\% & 8.7\% $\pm$ 1.3\% & \underline{\textbf{5.29\% $\pm$ 1.48\%}} & 5.35\% $\pm$ 1.79\% & 5.57\% $\pm$ 1.65\% \\ \hline
Modulus (GPa) SRJT & RMSE & 19.724 $\pm$ 3.744 & 16.760 $\pm$ 1.407 & 16.616 $\pm$ 2.231 & 18.810 $\pm$ 2.034 & \underline{\textbf{12.5840 $\pm$ 4.0755}} & 12.7843 $\pm$ 4.0212 & 13.1442 $\pm$ 2.6331 \\ 
 & MAE & 14.734 $\pm$ 2.604 & 13.085 $\pm$ 0.972 & 13.036 $\pm$ 1.861 & 14.438 $\pm$ 1.283 & \underline{\textbf{9.5227 $\pm$ 3.3418}} & 9.5866 $\pm$ 3.4949 & 10.0754 $\pm$ 2.3725  \\ 
 & R2 & -0.031 $\pm$ 0.427 & 0.283 $\pm$ 0.096 & 0.300 $\pm$ 0.116 & 0.124 $\pm$ 0.037 & 0.5064 $\pm$ 0.5120 & 0.4960 $\pm$ 0.5135 & \underline{\textbf{0.5173 $\pm$ 0.3643}}  \\ 
 & Spearman & 0.588 $\pm$ 0.118 & 0.613 $\pm$ 0.060 & 0.658 $\pm$ 0.078 & 0.478 $\pm$ 0.048 & \underline{\textbf{0.7525 $\pm$ 0.2478}} & 0.7505 $\pm$ 0.2624 & 0.7336 $\pm$ 0.2209 \\ 
 & SMAPE & 7.5\% $\pm$ 1.3\% & 6.7\% $\pm$ 0.4\% & 6.8\% $\pm$ 0.9\% & 7.4\% $\pm$ 0.6\% & \underline{\textbf{4.91\% $\pm$ 1.74\%}}&  4.93\% $\pm$ 1.82\% & 5.20\% $\pm$ 1.26\% \\ \hline
Avg HDYN/HQS & RMSE & 0.053 $\pm$ 0.004 & 0.047 $\pm$ 0.004 & 0.042 $\pm$ 0.008 & 0.041 $\pm$ 0.005 & 0.0262 $\pm$ 0.0122 & \underline{\textbf{0.0252 $\pm$ 0.0123}} & 0.0290 $\pm$ 0.0114 \\ 
 & MAE & 0.045 $\pm$ 0.002 & 0.038 $\pm$ 0.004 & 0.034 $\pm$ 0.007 & 0.033 $\pm$ 0.006 & 0.0206 $\pm$ 0.0095 & \underline{\textbf{0.0199 $\pm$ 0.0092}} & 0.0232 $\pm$ 0.0089 \\ 
 & R2 & -1.375 $\pm$ 1.100 & -0.760 $\pm$ 0.353 & -0.357 $\pm$ 0.196 & 0.043 $\pm$ 0.105 & 0.3371 $\pm$ 0.6079 & \underline{\textbf{0.3772 $\pm$ 0.5885}} & 0.3023 $\pm$ 0.4916 \\ 
 & Spearman & 0.267 $\pm$ 0.260 & 0.227 $\pm$ 0.309 & 0.227 $\pm$ 0.283 & 0.414 $\pm$ 0.145 & 0.4561 $\pm$ 0.5550 & \underline{\textbf{0.4585 $\pm$ 0.5749}} & 0.3726 $\pm$ 0.5455 \\ 
 & SMAPE & 3.9\% $\pm$ 0.2\% & 3.3\% $\pm$ 0.4\% & 2.9\% $\pm$ 0.6\% & 2.9\% $\pm$ 0.5\% & 1.80\% $\pm$ 0.82\% & \underline{\textbf{1.74\% $\pm$ 0.80\%}} & 2.03\% $\pm$ 0.78\% \\ \hline
Depth of Penetration (mm) FEM & RMSE & 0.179 $\pm$ 0.034 & 0.183 $\pm$ 0.054 & 0.154 $\pm$ 0.024 & 0.151 $\pm$ 0.039 & \underline{\textbf{0.1096 $\pm$ 0.0367}} &  0.1098 $\pm$ 0.0399 & 0.1261 $\pm$ 0.0394 \\ 
 & MAE & 0.126 $\pm$ 0.024 & 0.121 $\pm$ 0.026 & 0.110 $\pm$ 0.013 & 0.113 $\pm$ 0.033 & 0.0800 $\pm$ 0.0240 & \underline{\textbf{0.0788 $\pm$ 0.0280}} & 0.0954 $\pm$ 0.0287 \\ 
 & R2 & 0.460 $\pm$ 0.273 & 0.526 $\pm$ 0.188 & 0.632 $\pm$ 0.167 & 0.587 $\pm$ 0.161 & \underline{\textbf{0.6981 $\pm$ 0.2381}} & 0.6264 $\pm$ 0.2331 & 0.5967 $\pm$ 0.3180 \\ 
 & Spearman & 0.670 $\pm$ 0.167 & 0.678 $\pm$ 0.083 & 0.6770 $\pm$ 0.2993 & 0.742 $\pm$ 0.114 & \underline{\textbf{0.8649 $\pm$ 0.0619}} & 0.8563 $\pm$ 0.1350 & 0.8358 $\pm$ 0.1374 \\ 
 & SMAPE & 4.4\% $\pm$ 0.8\% & 4.2\% $\pm$ 1.0\% & 3.8\% $\pm$ 0.5\% & 3.9\% $\pm$ 1.1\% & 2.77\% $\pm$ 0.93\% & \underline{\textbf{2.73\% $\pm$ 1.03\%}} & 3.28\% $\pm$ 1.05\% \\ \hline
\end{tabular}%
}
\end{table}

The comparative analysis across surrogate models, as detailed in Tables \ref{tab:consolidated_metrics_1} and \ref{tab:consolidated_metrics_2}, provides valuable insights into the efficacy and suitability of different modeling strategies within high-entropy alloy (HEA) surrogate modeling. Among the evaluated models—XGBoost, cGP, Deep Gaussian Processes (DGP) variations (HDGP NP-All, HDGP NP-Main, HDGP P-All, HDGP P-Main), and the encoder-decoder neural network—the HDGP models incorporating prior knowledge (HDGP P-Main, HDGP P-All) consistently exhibited superior performance across the majority of tasks of primary experimental interest.

\begin{figure}[htbp]
  \centering
  \begin{subfigure}[t]{0.8\textwidth}
    \centering
    \begin{overpic}[width=\linewidth]{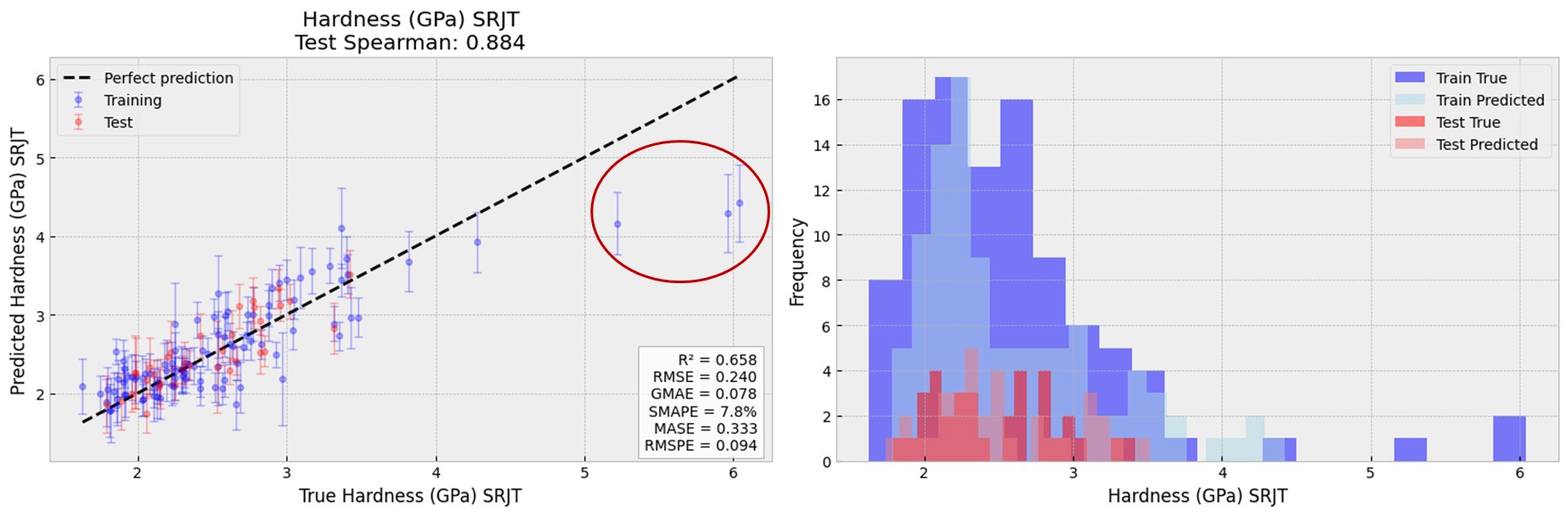}
      \put(0,31){\large\bfseries (a)}
    \end{overpic}
  \end{subfigure}
  \\[1em]
  \begin{subfigure}[t]{0.8\textwidth}
    \centering
    \begin{overpic}[width=\linewidth]{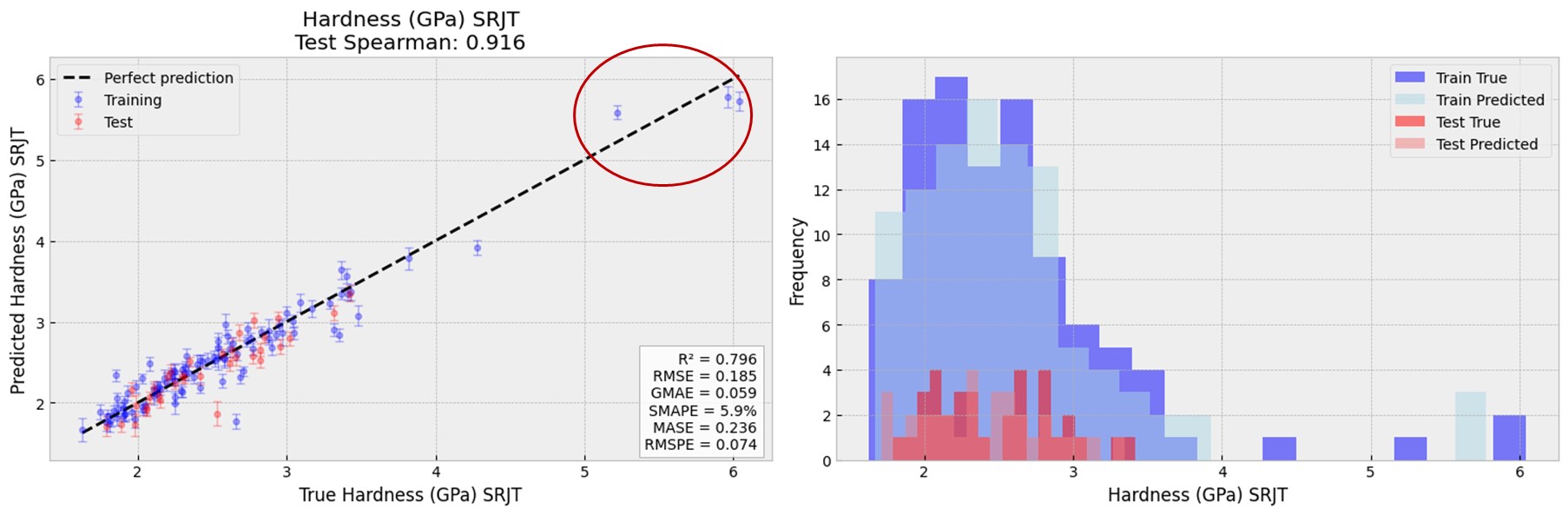}
      \put(0,31){\large\bfseries (b)}
    \end{overpic}
  \end{subfigure}

  \caption{%
    (a) Without prior injection the points within the red circle are further from actual value.  
    (b) With prior injection the points within the red circle tend to be closer to the actual value.
  }
  \label{fig:prior-comparison}
\end{figure}

Specifically, the HDGP-P-Main and HDGP-P-All model showed marked improvements in prediction accuracy, reflected by the lowest RMSE and SMAPE values, alongside the highest \(R^2\) and Spearman rank correlation coefficients for critical experimental properties such as Yield Strength, Ultimate Tensile Strength (UTS), elongation, hardness, and average hardness under dynamic/quasi-static conditions. Both of these models have mostly similar performance. For depth of penetration, modulus, hardness, elongation, yield strength and SFE, HDGP-P-All performed best. For UTS/YS, UTS and avg HDYN/HQS HDGP-P-Main performed better.  It is to mention that HDGP-P-All had encoder-decoder priors for all tasks except Varveillien yield strength, SFE and VEC.  The HDGP-P-All surpassed the performance of HDGP-P-Main in most of the tasks and it can possibly be attributed to the addition of auxiliary tasks, Auxiliary task may help the DGP generalize better by giving information about the correleation between tasks\cite{alvi2025hierarchical}. The exceptional performance of HDGP P-All and HDGP-P-Main over non-prior models can be primarily attributed to  the integration of encoder-decoder model as informative priors in the training process \cite{attari2024decoding}. Any GP-based model without prior defaults to constant mean in unseen region. Addition of prior makes the model converge to the prior at a specific input compositions rather than one constant value at all compositions\cite{vela2023data}. This makes the predictions more robust for out-of-distribution points. It also helps in leveraging expert knowledge or other parametric ML models, like encoder-decoder in our case. This is also illustrated in \ref{fig:prior-comparison} for the case of hardness in HDGP-NP-All and HDGP-P-All. For points inside the red circle, in the model with prior(b), the predictions don't default to a constant value in far-away regions. Thus, they are close to the actual values. Whereas in the model without prior(a), the mean tend to move toward a constant value, making the predictions defer from the actual values.

Interestingly,The performance of prior based models was followed by encoder-decoder, HDGP-NP-Main and HDGP-NP-All having intermediate performance. They performed better than XGBoost, cGP. But, performed worse than DGP models having priors. In between these two DGPs, HDGP-NP-All was the better one for most of the tasks. It had Varveillien yield strength, SFE and VEC as auxiliary tasks. Due to the hierarchical and deep structure of DGPs, the HDGP model effectively leveraged these auxiliary tasks, exploiting inter-task correlations to enhance predictive accuracy for main tasks compared to HDGP-NP-Main. This multi-task learning capability, intrinsic to DGP frameworks, allows the model to extract valuable latent information from auxiliary properties, thus significantly improving prediction robustness in sparse and noisy experimental settings \cite{damianou2013deep, salimbeni2017doubly}. Encoder-decoder model performed better than  HDGP-NP-Main and HDGP-NP-All, but it doesn't have the capacity of uncertainty quantification, so may not be useful in materials design or discovery campaign on its own. Moreover, the encoder-decoder neural network model, while theoretically capable of capturing complex nonlinearities, exhibited poorer generalization performance compared to DGPs with prior. This underperformance likely stems from its deterministic and parametric nature, sensitivity to hyperparameter settings, and susceptibility to overfitting when trained on smaller datasets \cite{ban2023effective}. Also, the encoder-decoder model had much larger number parameters, making it prone to overfitting for a small dataset like ours.

Contrastingly, the cGP and XGBoost models demonstrated several limitations despite their frequent use in regression tasks. Although XGBoost delivered competitive performance in specific auxiliary tasks (e.g., VarvYS), it notably lacked the flexibility and uncertainty quantification capabilities inherent to Gaussian Process-based methods. XGBoost, being a parametric, tree-based ensemble model, can easily overfit limited datasets, reducing its generalizability in complex materials design problems \cite{chen2016xgboost}. Similarly, cGP, despite their nonparametric Bayesian framework and robust uncertainty quantification, failed to match the performance of HDGP variants due to their inability to explicitly model inter-task correlations and input-dependent uncertainty through hierarchical structures. cGP assumes independence among outputs, limiting its ability to transfer knowledge effectively across related properties \cite{rasmussen2006gaussian}.

This trend in performance is also visible in train set metrics as attached in the supplementary file. Although XGBoost had the best training metric mostly, it was very closely followed by HDGP-P-All and HDGP-P-Main. These two DGP models performed better than cGP and encoder-decoder model in fitting the train data as well, affirming their superior generalization capability.

In summary, HDGP P-All emerges as the optimal modeling choice for predicting correlated material properties within HEA datasets, efficiently addressing common challenges of sparsity, noise, and missing data. Its hierarchical, Bayesian nature ensures robust uncertainty quantification and superior predictive performance, making it particularly suitable for guiding experimental design in materials informatics. It utilizes auxiliary tasks to generalize better. Also, It leverages the knowledge of encoder-decoder model as prior to perform better, providing promise as a hybrid model capable of uncertainty estimation and better generalization.  Moreover, figure \ref{fig:alltasks_grid} illustrates the parity plots and relevant test metrics for one specific train-test split in case of HDGP-P-All model, showing excellent accuracy. It is to be noted that the error bars in the plots are representative of the standard deviation as predicted by the model. The parity plots for all the other models and tables of train metrics are provided in the \textbf{supplementary file}.

\section{Discussion}

In this work, we systematically evaluated several surrogate modeling techniques, including conventional Gaussian processes (cGP), deep Gaussian processes (DGP), encoder-decoder neural networks, and XGBoost, to predict multiple correlated material properties in the complex AlCoCrCuFeMnNiV high entropy alloy (HEA) system. Our analysis demonstrated that the hierarchical DGP model, which incorporates informative priors and is trained across both main and auxiliary tasks (HDGP P-All), consistently outperformed other models in key mechanical properties, such as yield strength, ultimate tensile strength, elongation, hardness, modulus, and dynamic / quasistatic hardness ratios.

The superior performance of HDGP P-All can be primarily attributed to its hierarchical architecture and effective leverage of inter-task correlations, as well as the strategic integration of prior knowledge derived from a pre-trained encoder-decoder model. This combination enabled the model to robustly address common challenges inherent in materials informatics datasets, including heteroscedastic uncertainties, heterotopic observations, and missing data. In contrast, conventional GP, XGBoost, and deterministic encoder-decoder networks exhibited notable limitations, primarily due to their inability to explicitly quantify uncertainties, susceptibility to overfitting, and limited capacity for modeling correlations among multiple outputs.

In general, our findings strongly advocate the adoption of advanced hierarchical surrogate models with uncertainty-awareness such as DGP in materials discovery and optimization campaigns. These models provide substantial advantages by effectively capturing complex property interdependencies, robustly managing uncertainty, and maximizing predictive accuracy even with sparse and heterogeneous data sets. Future research should further optimize the integration of domain-specific prior knowledge and enhance computational efficiency to facilitate broader applications of these advanced methods in accelerated materials development.

\section{Methods}

\subsection{Conventional Gaussian Processes (cGP)}
cGPs provide a flexible Bayesian framework widely used for regression tasks due to their ability to explicitly quantify predictive uncertainty \cite{rasmussen2006gaussian}. Given a set of observations \(\mathcal{D} = \{\mathbf{X}_N, \mathbf{y}_N\}\), where \(\mathbf{X}_N = (\mathbf{x}_1, \mathbf{x}_2, \dots, \mathbf{x}_N)\) are the input points and \(\mathbf{y}_N = (y_1, y_2, \dots, y_N)\) are their corresponding outputs, a GP is fully defined by its mean function \(\mu(\mathbf{x})\) and the covariance function \(k(\mathbf{x}, \mathbf{x}')\). For an unseen input point \(\mathbf{x}^*\), the GP posterior prediction is given by:
\begin{equation}
p(f(\mathbf{x}^*)|\mathbf{X}_N, \mathbf{y}_N) = \mathcal{N}(\mu(\mathbf{x}^*), \sigma^2(\mathbf{x}^*))
\end{equation}
with predictive mean and variance:
\begin{align}
\mu(\mathbf{x}^*) &= \mathbf{k}(\mathbf{x}^*, \mathbf{X}_N)^\top\Big(\mathbf{K}(\mathbf{X}_N, \mathbf{X}_N) + \sigma^2_n\mathbf{I}\Big)^{-1}\mathbf{y}_N,\\[1ex]
\sigma^2(\mathbf{x}^*) &= k(\mathbf{x}^*, \mathbf{x}^*) - \mathbf{k}(\mathbf{x}^*, \mathbf{X}_N)^\top\Big(\mathbf{K}(\mathbf{X}_N, \mathbf{X}_N) + \sigma^2_n\mathbf{I}\Big)^{-1}\mathbf{k}(\mathbf{x}^*, \mathbf{X}_N)
\end{align}
Here, \(\mathbf{K}(\mathbf{X}_N, \mathbf{X}_N)\) denotes the covariance matrix computed from the kernel \(k\) for the training inputs, \(\mathbf{k}(\mathbf{x}^*, \mathbf{X}_N)\) is the covariance vector between the test point \(\mathbf{x}^*\) and the training inputs, and \(\sigma^2_n\) represents the observation noise variance. A commonly adopted kernel is the squared exponential (RBF) kernel:
\begin{equation}
k(\mathbf{x}, \mathbf{x}') = \exp\left(-\frac{||\mathbf{x}-\mathbf{x}'||^2}{2\ell^2}\right)
\end{equation}
which characterizes the similarity between input points based on their Euclidean distance, with \(\ell\) controlling the length scale. This description of conventional GPs lays the foundational groundwork for understanding more complex modeling approaches.

\subsection{Deep Gaussian Processes (DGP)}
Building directly on the fundamental concepts introduced above, Deep Gaussian Processes (DGPs) extend conventional GPs by incorporating a hierarchical structure. This hierarchical arrangement enables the modeling of more intricate, non-linear relationships and input-dependent uncertainties, thereby addressing challenges that arise in complex datasets \cite{damianou2013deep,salimbeni2017doubly}. In a DGP with \(L\) hidden layers, the generative model is described by:
\begin{align}
\mathbf{y}_n &= f^{(L)}(\mathbf{h}^{(L-1)}_n) + \epsilon_n^{(L)}, \quad \epsilon_n^{(L)} \sim \mathcal{N}(0, \sigma_L^2 I),\\[1ex]
\mathbf{h}_n^{(L-1)} &= f^{(L-1)}(\mathbf{h}_n^{(L-2)}) + \epsilon_n^{(L-1)}, \quad \epsilon_n^{(L-1)} \sim \mathcal{N}(0, \sigma_{L-1}^2 I),\\[1ex]
&\vdots\\[1ex]
\mathbf{h}_n^{(1)} &= f^{(1)}(\mathbf{x}_n) + \epsilon_n^{(1)}, \quad \epsilon_n^{(1)} \sim \mathcal{N}(0, \sigma_1^2 I)
\end{align}
Each function \(f^{(i)}\) is modeled as a GP with its own covariance structure, effectively capturing different levels of abstraction from the input data. Given the increased computational complexity of DGPs, training often involves variational inference techniques. Specifically, the model is optimized by maximizing the Evidence Lower Bound (ELBO):
\begin{equation}
\mathcal{L} = \int Q \log \frac{p(\mathbf{Y}, \mathbf{H}, \mathbf{F})}{Q},
\end{equation}
where \(Q\) represents the variational distribution that approximates the true posterior. Gradient-based optimization methods are typically employed to balance model fit and complexity \cite{salimbeni2017doubly,hensman2013gaussian}. This progression from conventional to deep GP structures demonstrates the evolution from simple probabilistic modeling to a framework capable of handling greater nonlinearity and uncertainty.

\subsection{Isotopic vs. Heterotopic Data}
Continuing from the dataset description, we now address a key data structuring concern relevant to the modeling techniques presented earlier. In multi-output Gaussian process modeling, distinguishing between isotopic and heterotopic data is essential due to its implications on model performance and uncertainty quantification \cite{alvarez2012kernels}. 

Isotopic data, where all tasks are observed at the same set of input points, provides a straightforward scenario:
\begin{equation}
\mathcal{D}_{\text{iso}} = \{(\mathbf{x}_i, f_1(\mathbf{x}_i), \dots, f_T(\mathbf{x}_i))\}_{i=1}^{N},
\end{equation}
where \(T\) is the number of tasks. Conversely, heterotopic data arises when tasks are observed at different input points:
\begin{equation}
\mathcal{D}_{\text{hetero}} = \bigcup_{t=1}^T \{(\mathbf{x}_{i,t}, f_t(\mathbf{x}_{i,t}))\}_{i=1}^{N_t},
\end{equation}
leading to sparsely and partially missing observations for some tasks. This characteristic is particularly common in materials informatics due to the varying feasibility or cost of data collection for different properties \cite{khatamsaz2021efficiently,ghoreishi2019efficient}. 

By incorporating the capability to handle heterotopic data, DGP models offer a significant advantage. Their ability to leverage inter-task correlations not only maximizes information utility from incomplete datasets but also enhances predictive accuracy and uncertainty quantification---key factors for accelerating materials discovery. The clear delineation between isotopic and heterotopic data thus bridges the theoretical modeling framework with the practical challenges encountered in experimental datasets.

The DGP models implemented in this work consist of two-layered variational Gaussian processes, each configured with 10 latent GPs and implemented using BoTorch \cite{balandat2020botorch}. To ensure numerical stability and consistent predictions, all input and output variables were scaled prior to model fitting and subsequently descaled to original units during result visualization.

\begin{figure}[!ht]
    \centering
    \includegraphics[width=0.85\textwidth]{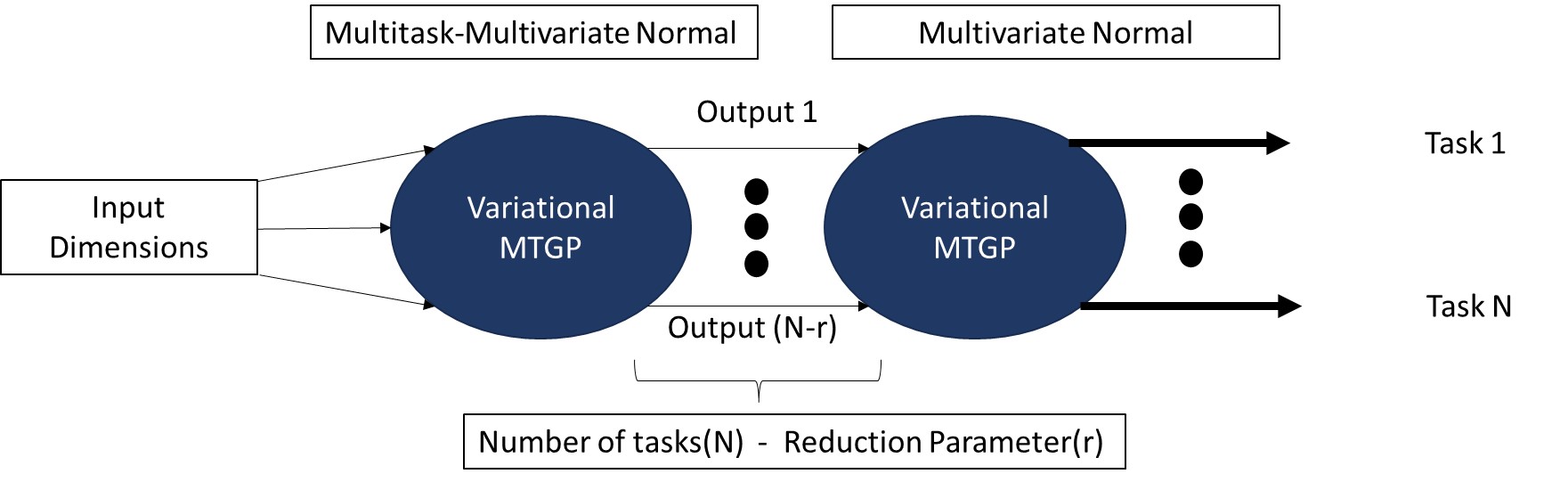}
    \caption{Schematic representation of the two-layered variational Deep Gaussian Process (DGP) architecture. Each variational Multi-task Gaussian Process (MTGP) layer reduces dimensionality from input to latent representation, governed by the reduction parameter.}
    \label{fig:dgp_architecture}
\end{figure}

A critical hyperparameter specific to the DGP architectures evaluated is the reduction parameter, illustrated schematically in Figure~\ref{fig:dgp_architecture}. The reduction parameter dictates the dimensionality reduction performed between the first and second layers of the DGP, calculated as the number of tasks minus a chosen integer. Selecting an optimal reduction parameter is essential, as it influences the capacity of the DGP to capture inter-task correlations and the complexity of latent representations. We systematically explored different values of this hyperparameter by fitting models with varying reduction parameters and selected the best-performing value based on cross-validation accuracy metrics. We plotted the spearman rank coeeficient and RMSE over different reduction parameters and choose the reduction parameter that yielded highest spearman rank co-efficient and lowest RMSE value.

\subsection{Encoder-Decoder Model for Tabular Data Learning}

The encoder-decoder framework is a powerful approach for learning complex input-output relationships in regression tasks and has shown competitive performance compared to tree-based methods on small datasets~\cite{attari2024decoding}. Its performance improves with larger datasets, enabling the model to capture more nuanced patterns and generalize better to unseen inputs. Although the BIRDSHOT dataset is relatively small, we used the encoder-decoder model to benchmark its performance against deep Gaussian processes, as both are capable of modeling non-linear mappings and uncertainty in high-dimensional spaces. Training involves minimizing the difference between the predicted and actual values using a loss function, typically optimized with gradient-based methods. After training, the model can generalize to new inputs and produce corresponding predictions. Although the model offers strong predictive performance, its internal representations are often not interpretable. However, explainability can be improved using attention mechanisms, feature attribution methods, and visualization tools. Different neural architectures including but not limited to dense networks, Disjunctive Normal Form Networks (DNF-Nets), and Convolutional Neural Networks (CNNs) can be used at the core of encoder and decoder. These architectures can enhance accuracy by capturing detailed patterns and filtering noise. In this study, we used regularized dense networks, also known as feedforward networks. Non-linear activation functions like ReLU or sigmoid introduced flexibility and regularization methods such as L2 penalties used to reduce overfitting. 

%

\section{Data Availability}
The data supporting the results of this study can be found in the following Github repository. 

\section{Code Availability}
The code and data supporting the results of this study can be found in the following github repository as a google colaboratory notebook. The encoder-decoder model for tabular learning and regression tasks can be found in \url{https://github.com/vahid2364/DataScribe_DeepTabularLearning}
The other models can be found in
\url{https://github.com/sheikhahnaf/DGP_with_EncoderDecoderPrior}

\section{Acknowledgements}
This material is based on work supported by the Texas A\&M University System National Laboratories Office of the Texas A\&M University System and Los Alamos National Laboratory as part of the Joint Research Collaboration Program. Any opinions, findings, conclusions or recommendations expressed in this material are those of the author(s) and do not necessarily reflect the views of the Los Alamos National Laboratory or The Texas A\&M University System. JJ acknowledges support from the Los Alamos National Laboratory Laboratory (LANL) Laboratory Directed Research and Development Program under project number 20220815PRD4.
DP acknowledges support from the Los Alamos National Laboratory Laboratory (LANL) Laboratory Directed Research and Development Program under project number 20220063DR. LANL is operated by Triad National Security, LLC, for the National Nuclear Security Administration of U.S. Department of Energy (Contract No. 89233218CNA000001). Original data were generated within the BIRDSHOT Center (https://birdshot.tamu.edu), supported by the Army Research Laboratory under Cooperative Agreement (CA) Number
W911NF-22-2-0106 (MM, DK, DA, VA and RA acknowledge partial support from this CA). NF acknowledges support from AFRL through a subcontract with ARCTOS, TOPS VI (165852-19F5830-19-02-C1).The authors acknowledge the support from the U.S. Department of Energy (DOE) ARPA-E CHADWICK Program through Project DE‐AR0001988. Calculations were carried out at Texas A\&M High-Performance Research Computing (HPRC).


\bibliographystyle{elsarticle-num}

\bibliography{main}

\end{document}